%% file: root.tex
\documentclass[letterpaper, 10 pt, conference]{ieeeconf}  

\IEEEoverridecommandlockouts                              

\usepackage{graphics} 
\usepackage{graphicx}
\usepackage{epsfig} 
\usepackage{cite}
\usepackage{comment}
\usepackage{soul}
\usepackage{adjustbox}
\usepackage{multirow}
\usepackage{subcaption}
\usepackage{graphicx}
\usepackage{textcomp}
\usepackage{wrapfig}
\usepackage{siunitx}
\usepackage{multirow}
\usepackage[table,xcdraw]{xcolor}
\usepackage{tikz}
\usepackage{wrapfig}
\usepackage{lscape}
\usepackage{url}
\usepackage{rotating}
\usepackage{float}
\renewcommand{\baselinestretch}{0.91}

\usepackage[hidelinks]{hyperref}
\usepackage{cleveref} 

\usetikzlibrary{automata,arrows,positioning,calc,shapes,patterns,patterns.meta,arrows.meta}
\input{Tex/Comments}
\input{Images/control_diagram_helper}
\DeclareUnicodeCharacter{0308}{HERE!HERE!}

\title{\LARGE \bf
OPENGRASP-LITE Version 1.0: A Tactile Artificial Hand with a Compliant Linkage Mechanism
}

\author{Sonja Gro{\ss}$^{*,1,2}$, Michael Ratzel$^{*,1}$, Edgar Welte$^{1}$, Diego Hidalgo-Carvajal$^{1,2}$, Lingyun Chen$^{1,2}$,\\ Edmundo Pozo Fortuni\'c$^{1}$,  Amartya Ganguly$^{1}$, Abdalla Swikir$^{1,2}$, and Sami Haddadin$^{1,2}$
\\ 
\thanks{$^{*}$ equal contribution}
\thanks{This work was funded by the German
Research Foundation (DFG, Deutsche Forschungsgemein-
schaft) as part of Germany’s Excellence Strategy – EXC
2050/1 – Project ID 390696704 – Cluster of Excellence
“Centre for Tactile Internet with Human-in-the-Loop” (CeTI) of Technische University of Dresden. This work was also supported by the Federal Ministry of Education and Research of the Federal Republic of Germany (BMBF) by funding the project AI.D under the Project Number 16ME0539K.}
\thanks{$^{1}$ {\small \{lingyun.chen, sonja.gross, diego.hidalgo-carvajal, edgar.welte, michael.ratzel, amartya.ganguly, abdalla.swikir, haddadin\}@tum.de}, 
Technical University of Munich, Germany; TUM School of Computation, Information and Technology (CIT); Chair of Robotics and Systems Intelligence (RSI); Munich Institute of Robotics and Machine Intelligence (MIRMI)}
\thanks{$^{2}$ the Centre for Tactile Internet with Human-in-the-Loop (CeTI)
}
}

\begin{document}

\maketitle
\thispagestyle{empty}
\pagestyle{empty}


\input{Sections/Abstract}
\input{Sections/Introduction}
\input{Sections/Methods}
\input{Sections/Results}
\input{Sections/Discussion_and_conclusion}

\addtolength{\textheight}{0.5cm}   





\bibliographystyle{IEEEtran}
\bibliography{root.bib}

\end{document}

%% file: Tex/Comments.tex
\definecolor{ag}{rgb}{0.08, 0.33, 0.9}

\definecolor{mr}{rgb}{0.4,0.7,0.1}

\definecolor{sg}{rgb}{0.5,0.1,0.5}

\definecolor{lc}{rgb}{0.5,0.1,0.9}

\definecolor{ew}{rgb}{0.63,0.68,0} 

\definecolor{dhc}{rgb}{0.1,0.7,0.9}

\definecolor{todo}{rgb}{1.0, 0., 0.}

\colorlet{as}{magenta}

%% file: Images/control_diagram_helper.tex
\newsavebox{\SaveBoxFilterLimit}
\savebox{\SaveBoxFilterLimit}{%
\begin{tikzpicture}
    \draw[draw opacity=0] (-0.5,-0.5) -- (0.5,0.5);
    \draw[thick] (-0.4,-0.4) -- (-0.2,-0.4) -- (0.2,0.4) -- (0.4,0.4);
\end{tikzpicture}%
}
\newsavebox{\SaveBoxFilterSign}
\savebox{\SaveBoxFilterSign}{%
\begin{tikzpicture}
    \draw[draw opacity=0] (-0.5,-0.5) -- (0.5,0.5);
    \draw[thick] (-0.4,-0.4) -- (0,-0.4) -- (0,0.4) -- (0.4,0.4);
    \node[anchor=west,inner sep=0] at (-0.35,-0.15) {\textsf{-1}};
    \node[anchor=east,inner sep=0] at (0.35,0.15) {\textsf{1}};
\end{tikzpicture}%
}

\newsavebox{\SaveBoxFilterEMA}
\savebox{\SaveBoxFilterEMA}{%
\begin{tikzpicture}
    \draw[draw opacity=0] (-0.5,-0.5) -- (0.5,0.5);
    \draw[thick] (-0.4,0.35) -- (-0.1,0.35) .. controls (0.15,0.35) .. (0.4, -0.1);

    \node[anchor=south,inner sep=0] at (0,-0.4) {\textsf{EMA}};
\end{tikzpicture}%
}

\newcommand{\filterLimit}{\usebox{\SaveBoxFilterLimit}}
\newcommand{\filterSign}{\usebox{\SaveBoxFilterSign}}
\newcommand{\filterEMA}{\usebox{\SaveBoxFilterEMA}}

%% file: Sections/Abstract.tex
\begin{abstract}
Recent research has seen notable progress in the development of linkage-based artificial hands. While previous designs have focused on adaptive grasping, dexterity and biomimetic artificial skin, only a few systems have proposed a lightweight, accessible solution integrating tactile sensing with a compliant linkage-based mechanism. This paper introduces OPENGRASP LITE, an open-source, highly integrated, tactile, and lightweight artificial hand. Leveraging compliant linkage systems and MEMS barometer-based tactile sensing, it offers versatile grasping capabilities with six degrees of actuation. By providing tactile sensors and enabling soft grasping, it serves as an accessible platform for further research in tactile artificial hands.
\end{abstract}

%% file: Sections/Introduction.tex
\section{INTRODUCTION}
\label{section: introduction}
\input{Tables/soa}
The advancement of artificial hands in robotics and prosthetics has reached a pivotal moment, with the integration of advanced transmission mechanisms and tactile sensing technologies leading the charge. These innovations not only enhance the mechanical dexterity of artificial hands but also imbue them with a sense of touch, significantly improving interaction with the world \cite{haddadin2018tactile, Hogan2020, Donati2024}. 
Recent research has advanced the field of linkage-based artificial hands \cite{kashef2020robotic}. For instance, a design for prosthetic fingers was introduced with a compliant four-bar linkage to absorb impacts and prevent breakdowns \cite{choi2017compliant}.
The MCR-Hand III, a low-cost linkage-spring-tendon-integrated compliant anthropomorphic robotic hand presented a structured design and formulated the kinematics of the linkage and tendon-driven fingers \cite{yang2021low}.
More recently, a three-fingered robotic hand has been introduced in \cite{li2023linkage}. It used a human hand motion-inspired linkage-driven underactuated mechanism to perform adaptive grasping and in-hand manipulation, offering increased dexterity. 

Significant strides have also been made in low-cost and open-source robotic hand design. In \cite{ma2017yale}, the authors presented an initiative to advance the design and use of open-source robotic hands built through rapid prototyping techniques. The HRI hand is a low-cost, open-source anthropomorphic robot hand system developed as an end-effector for collaborative robot manipulators, which can be built using 3D printing and provides pre-shaping motion similar to the human hand \cite{park2020open}.  
\begin{figure}[tb]
    \centering    
    \includegraphics[width={0.48\textwidth}]{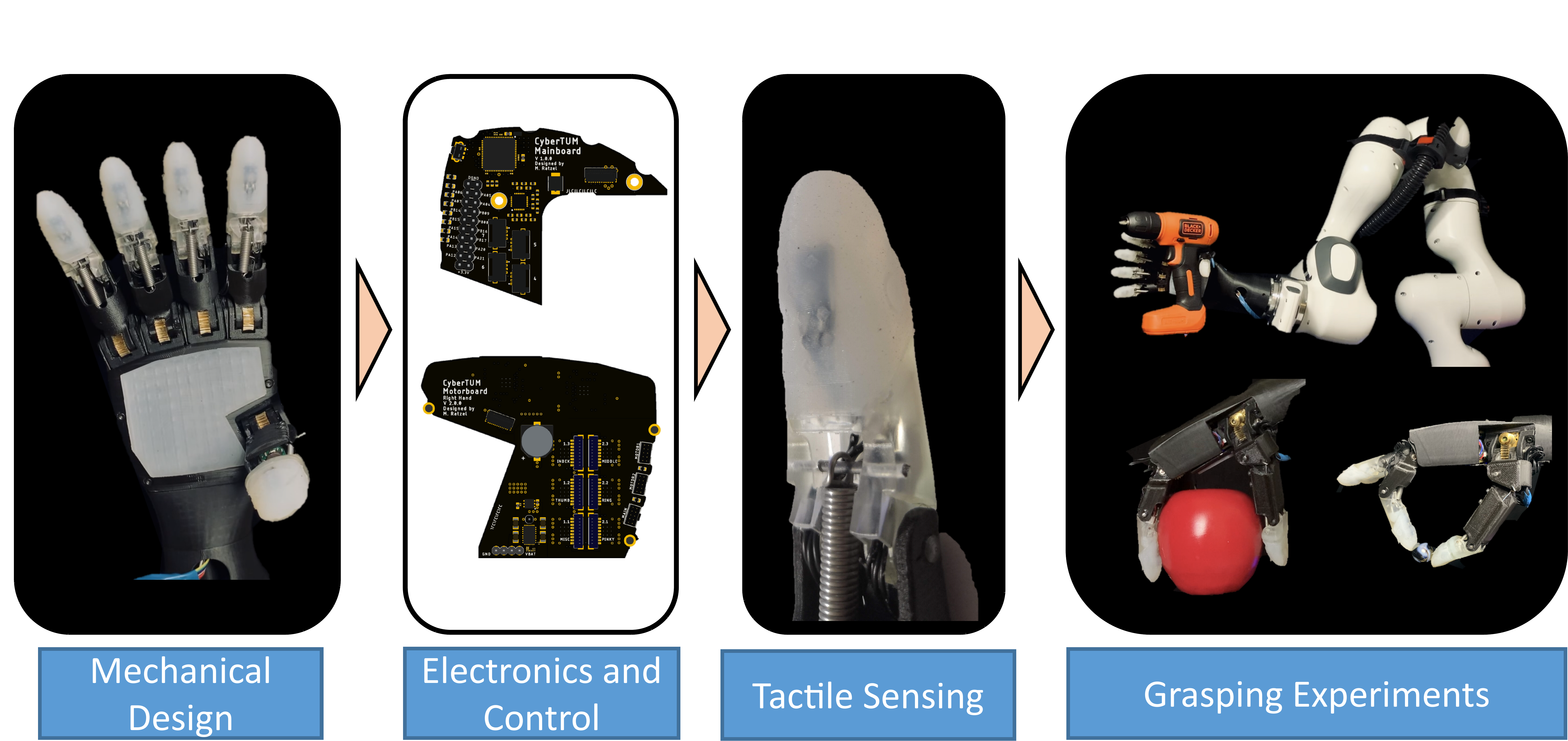}    
    \caption{Overview of this paper.}
    \label{fig: overview}    
    \vspace{-1.5\baselineskip}
\end{figure}
Moreover, current research on including tactile sensing strives for advanced control systems, that mimic human manipulation abilities. Soft, skin-like sensors remain a popular approach including (piezo)resistive, piezoelectric, capacitive, barometric, magnetic or optical transmission concepts \cite{Wang2023}. 
More recent research explored advanced sensor technologies such as MEMS-based (micro-electromechanical system) barometer arrays on the Shadow Dexterous Hand \cite{Koiva2020}, optical sensors integrated into a robotic hand for manipulation reinforcement learning\cite{khandate2023sampling}, or the D'Manus hand, which implemented large-area, immersive tactile sensing across its palm and fingertips\cite{bhirangi2023all}.  Although significant contributions were made in the mechatronics development of tactile artificial hands, only a few systems combine these into accessible, lightweight, and open-source systems as shown in \cref{tab: robotic_hand_comparison}.

Therefore, this paper introduces the design and implementation of a new highly integrated tactile, lightweight, cost-effective, and open-source linkage-based artificial hand called OPENGRASP LITE \footnote{\url{https://gitlab.com/tum-mirmi-handgroup/opengrasp-lite}}. We aim to provide a combination of cutting-edge technology by integrating the advantages of 1) a compliant linkage-based system,  2) the incorporation of MEMS-based fingertip sensors for tactile sensing and 3) the accessibility of an open-source platform. These key aspects would enable our system as a development platform  (white box) for the research community for a wide range of applications. With its 6 DoA (one per finger and two for thumb), embedded tactile sensor data combined with the motor control processes, precise closed-control loop applications can be achieved in robotic teleoperation, prosthetics or geriatronics.

%% file: Tables/soa.tex
\begin{table*}[tb]
\centering
\begin{tabular}{ccccccccc}
\hline
Hand             & Mechanism           & No. of Finger & DoF & DoA & Weight {[}\unit{\g}{]} & Tactile-capable & Approx. Cost {[}EUR{]} & Open-source \\ \hline
Choi et, al      & Linkage-based       & 5             & 6   & 6   & 342            & Yes             & $\sim$500                  & No          \\
MCR-Hand III     & Linkage-based       & 5             & 21  & 16  & \textless{}500 & Yes             & \textless{}750             & No          \\
ILDA             & Linkage-based       & 5             & 20  & 15  & 1100           & Yes             & \textgreater{}5000 *       & No          \\
Li et, al        & Linkage-based       & 3             & 12  & 6   & -              & No              & -                          & No          \\
OpenHand Model O & Tendon-based        & 3             & 7   & 4   & 752            & Yes             & $\sim$1500                 & Yes         \\
HRI              & Linkage-based       & 5             & 15  & 6   & 570            & No              & $\sim$460                  & Yes         \\
LEAP             & Motor-direct-driven & 4             & 16  & 16  & $\sim$500      & No              & $\sim$2000                 & Yes         \\
(Proposed)       & Linkage-based       & 5             & 11  & 6   & 370            & Yes             & $\sim$500                  & Yes         \\ \hline\end{tabular}
\caption{Robotic hand comparison.}
\label{tab: robotic_hand_comparison}
\vspace{-1.5\baselineskip}
\end{table*}

%% file: Sections/Methods.tex
\section{Design and Implementation}
\label{section: methods}
In this section, we present the essential principles of the system design and development, including its overall architecture and sub-modules. We provide a comprehensive description of the specific hardware components that were used, enabling anyone to replicate our design.
\subsection{Design principle and Architecture}
The design of OPENGRASP-Lite harnessed an anthropomorphic principle for its architecture and shape, as depicted in Fig.\ref{fig: handDesign}a. Its dimensions were $\qty{200}{\mm} \textrm{ (length)} \times \qty{86}{\mm} \textrm{ (width)} \times \qty{219}{\mm} \textrm{ (circumference)}$. It comprised of four fingers and a thumb, with one degree of actuation (DoA) for each finger while the thumb had two. All actuation units were placed within the hand palm chamber of $\sim50 cm^3$. Each unit comprised of a DC motor (Pololu, USA) with an embedded gearbox and magnetic quadrature encoder (4760, Pololu, USA), along with an attached worm gear (SCR05, Lemo-Solar, Germany). An embedded electronic controller was also housed within the palm using two custom-made PCBs for actuator control and data acquisition. The robot interface was at the wrist.

\subsection{Finger Design and Transmission}
\begin{figure*}[tb]
    \centering    
    \includegraphics[width={0.7\textwidth}]{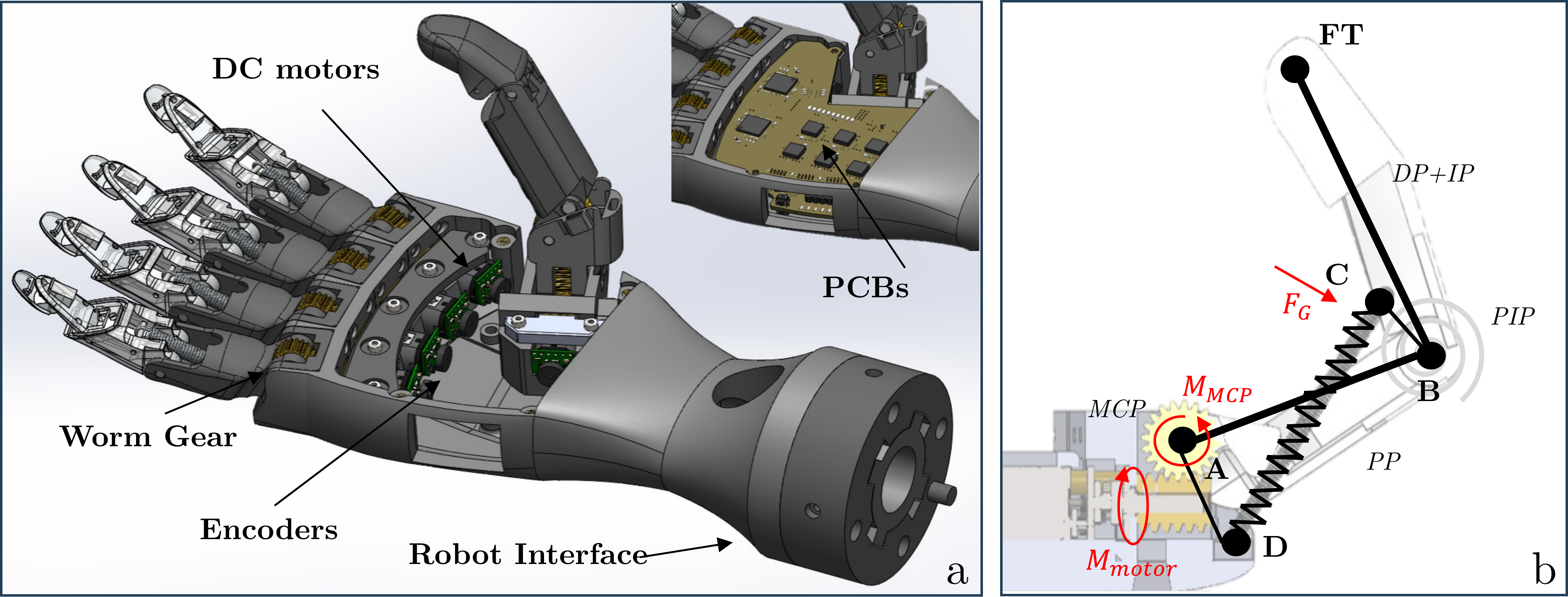}    
    \caption{Design of the hand. (a) 6 DOA driven by DC motors with attached worm gears and encoders, custom PCBs and, a robot interface. (b) Four-bar linkage mechanism for fingers consisting of three rigid and one compliant linkage.}
    \label{fig: handDesign} 
    \vspace{-1.5\baselineskip}
\end{figure*}

During the mechanical design phase, two key considerations were emphasized: prioritizing functionality and overall dexterity and minimizing the complexity of the transmission mechanism. As a result, a linkage-based transmission system was chosen for finger motion. Each finger comprised distal (DP), intermediate (IP), and proximal phalanges (PP) connected to their respective Distal Interphalangeal (DIP), Proximal Interphalangeal (PIP), and Metacarpophalangeal (MCP) joints. The DIP joint was set at a fixed angle of \ang{20}, informed by results from \cite{laffranchi2020hannes} to streamline movement across crucial degrees of freedom in the hand. This simplification facilitated the design of four-bar linkages for each finger, as depicted in \cref{fig: handDesign}b.

The linkage mechanism consisted of four revolute joints (A, B, C, D), where only A was connected to the actuation unit. These joints connected four linkages out of which three (AB, BC, and AD) were rigid, i.e. had fixed lengths, and one (CD) was an elastic-compliant element. Linkage AB was the input link as it was connected to the motor and drove the entire mechanism. In practice, this link was represented by the PP of the finger as seen in \cref{fig: handDesign}b. Linkage AD was the ground link and was fixed to the palm. Linkage BC (coupler) connected the compliant and input linkages. The compliant linkage (follower) connected joints C and D and reacted depending on the input motions.

The finger dimensions were determined by conducting a geometrical simulation of the finger motion based on different phalanges lengths using MATLAB (R2021a, MathWorks, Inc., United States). 
The linkage configuration was chosen for the ability to grasp objects used in daily living as described in \cite{Cybathlon}. The objects presented minimal and maximal radii of $\sim$\qty{15}{\mm} and $\sim$\qty{70}{\mm}, respectively. This evaluation lead to finger dimensions of \qty{36}{\mm} for the PP, \qty{23.5}{\mm} for the MP and \qty{21.00}{\mm} for the DP.

The compliant linkage-based finger design, with fixed DIP joints, offered distinct advantages. It simplified the mechanical structure, reducing costs and enhancing reliability, while still allowing for a broad range of grasping functions. This ensured the fingers were capable of handling various object sizes and shapes with ease, mirroring the dexterity of a human hand. Furthermore, the adaptability of the design allowed for the customization of individual applications. This innovative approach not only makes artificial hands more accessible but also sets the stage for incorporating advanced control technologies, enhancing user experience.

\subsection{Thumb Design and Actuation}

The thumb base was designed as a separate body, which was screwed to the palm. Its workspace enabled the most common grasping types, including precision grasps, such as pinch and tripod. Distinguishing itself from the fingers, the thumb featured two DOAs: one DOA facilitated radial abduction, while the second governed flexion and extension movements. The motor and worm gear responsible for abduction were situated within the palm, while the motor for flexion was integrated into the PP (refer to the attached repository for a detailed view).

\subsection{Motor Selection}
To determine the appropriate actuator combination for each finger, our objective was to accommodate a payload of at least \qty{2.75}{\kg}, analogous to the weight of a loaded water crate (\qty{5.5}{\kg}), based on \cite{Cybathlon} that can be lifted with two hands. Each finger was designed to support such a load. Based on this premise, the transmission ratios and motors were sized as follows:

\begin{equation}\label{eq: force_per_finger1}
      F_{G} = \qty{2.75}{\kg} \times \qty{9.81}{\metre\per\square\second} = \qty{27}{\N}.
\end{equation}
Under the assumption that the force will act on the PIP joint as depicted in \cref{fig: handDesign}b, the resulting MCP joint torque was
\begin{equation}\label{eq: force_per_finger2}
      M_\mathrm{MCP} = \qty{27}{\N} \times \qty{36}{\mm} = \qty{971}{\N\mm}.
\end{equation}
Due to the gear ratio of the worm gear attached to the motor, the minimal required motor torque was calculated as follows:
\begin{equation}\label{eq: force_per_finger3}
      M_{motor} = (1/20) \times \qty{971}{\N\mm} \approx \qty{49}{\N\mm}.
\end{equation}
Prioritizing low weight, minimal driver requirements, and high torque transmission, we chose brushed DC motors featuring a micro metal gear and encoder shaft (HPCB \qty{6}{\V} dual-shaft, Pololu, USA). For the little, ring, middle finger, and thumb abduction, we selected a gear ratio of 75:1, providing a stall torque of \qty{107}{\N\mm}, resulting in a safety factor of 2.2 to account for potential frictional errors arising from calculation simplifications. To further enhance safety, the thumb and index finger were equipped with a higher gear ratio of 100:1, given their increased involvement in grasping activities.

\subsection{Tactile Sensors}
\begin{figure*}[tb]
    \centering    
    \includegraphics[width=0.8\linewidth]{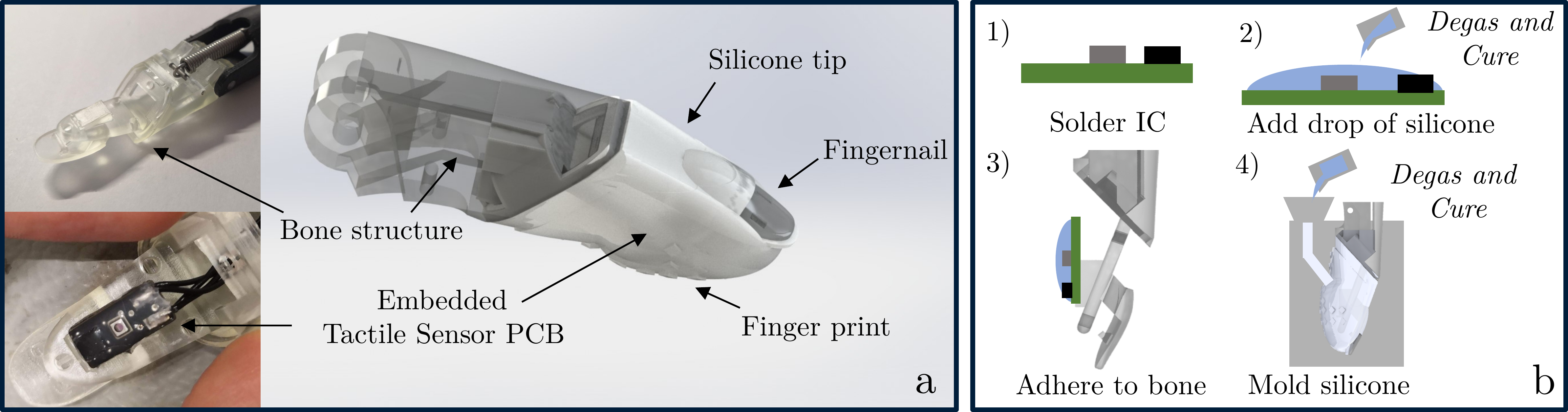}  \caption{a) Tactile Fingertip with a bone structure and tactile sensors encased in silicone. b) Sensor manufacturing.}
    \label{fig: fingertip B}
    \vspace{-0.5\baselineskip}
\end{figure*}

To facilitate normal force sensing through a single taxel, the tactile sensors were embedded in the PP of each finger, utilizing a concept originally introduced by \cite{tenzer2012inexpensive}. As depicted in \cref{fig: fingertip B}a, the sensors consisted of a custom-designed miniature sensor PCB encased into a silicone fingertip (DragonSkin 10, Smooth-On Inc., USA). The sensor PCB held a MEMS-based barometer sensor integrated circuit (IC) that utilized a pressure-sensitive membrane (BMP384, Bosch, Germany). Due to the silicone encasing, force application on the silicone leads to the deformation of the membrane, and therefore, a sensor signal change. 
For manufacturing as depicted in \cref{fig: fingertip B}b, we first soldered the ICs to the PCB and covered the barometer with a drop of silicone. Then, we degassed it in a vacuum chamber (Vacuum system with pump VP1200 and Thermo Scientific Nalgene vacuum chamber, Silikonfabrik, Germany) to remove any air between the silicone and sensitive membrane. Subsequently, we glued the PCB to the bone and cast moulded the silicone tip around it using a custom-built casting mould. Multiple encasements facilitated a mechanical bonding between the bone and silicone, obviating the necessity for adhesives.
For the sensor read-out via the Inter-Integrated Circuit (I$^2$C) protocol, we employed the manufacturer's library with the following parameters: temperature compensation enabled, pressure and temperature oversampling rate set to 2, reading frequency of \qty{100}{\Hz}, IIR filter constant of 3.
\begin{figure*}[tb]
    \centering
    \begin{subfigure}[b]{0.469\linewidth}
        \centering    
        \newcommand{\electronicsOverviewScale}{0.539}
        \newcommand{\electronicsOverviewColor}{black!0}
        \input{Images/electronics_overview}
    \end{subfigure}
    \hfill
    \begin{subfigure}[b]{0.522\linewidth}
        \centering    
        \newcommand{\controlLoopScale}{0.539}
        \input{Images/control_loop}
    \end{subfigure}
    \caption{a) Architecture overview of the hand control electronics. b) Motor control block diagram: cascaded position, velocity, and current controller}
    \label{fig:electronics_overview}
    \label{fig:control_loop}
    \vspace{-1.5\baselineskip}
\end{figure*}
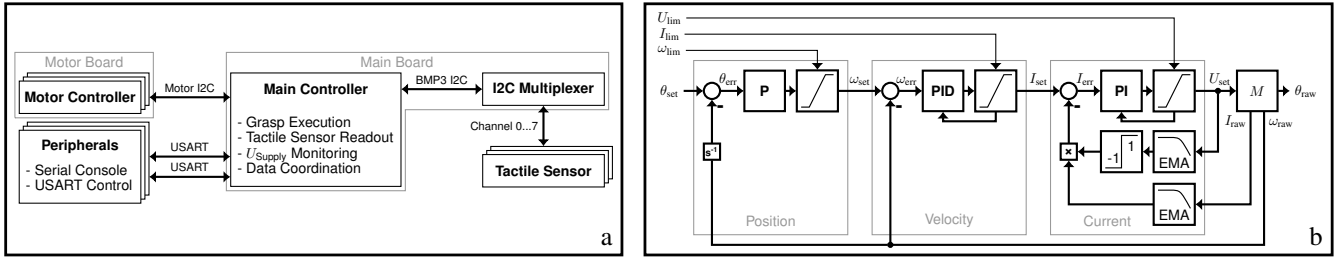

\subsection{Electronics Design and Software Architecture}
Given the space constraints of the design, two custom-made embedded PCBs were developed and installed inside the palm. The drive board was comprised of two Arm Cortex M4F @100MHz microcontroller (\unit{\micro C}) units (SAMDJ19A, Microchip Technology Inc., China), each connected to three single channel H-Bridge drivers (TB9051FTG, Toshiba Electronic Devices \& Storage Corporation, Japan), to drive three brushed DC motors using a low-level control architecture. The corresponding embedded position encoder sensors and current analogue signals were also measured by each \unit{\micro C}. The main board, based on the same class of processor, handled the task controller, data coordination and high-level grasp execution, as described in \cref{subsection: Grasp Execution}. An I$^2$C bus was used to communicate to the driver board at 50\,Hz; while an additional I$^2$2C bus measured the compatible tactile sensors at a 25\,Hz rate. Two Additional universal synchronous/asynchronous receiver/transmitter (USART) channels were used for debugging and task-based communication.
\subsection{Control approach}
Our control strategy comprised two layers, as depicted in \cref{fig:electronics_overview}a: a lower-level cascaded control running on the motor board, responsible for managing the position, velocity, and current of each motor, and a high-level grasp control operating on the main board, which facilitated the initiation of various pre-defined grasping types via the USART communication interface.

\subsubsection{Motor Control}
We utilized a cascaded multi-rate control loop for each motor as shown in \cref{fig:control_loop}b. The inner loop consisted of a PI-based current control running at 10 kHz. The middle loop was a PID-based velocity control running at 1 kHz. The outer loop utilized a P-based position control loop also running at 1 kHz. All integrators implemented an anti-windup function with configurable limits. The position and velocity control loops could be bypassed via software by the configuration of the task controller and directly setting the corresponding desired velocity $\omega_\text{set}$ or desired current $I_\text{set}$. As the current measurement functionality provided by the used motor driver only allowed measuring the absolute current flowing through the motor $M$, the measured current was multiplied by the sign of the set output voltage $U_\text{set}$. As the motor inductance prevented the current from immediately changing after switching the voltage, an exponential moving average (EMA) was applied to $U_\text{set}$ to emulate the behaviour of the motor LR-circuit. With the motor parameters of $R=\qty{3.2}{\ohm}$ and $L=\qty{0.6}{\mH}$ the step response reached \qty{99.3}{\percent} of the target value after $5\tau$ with $\tau=\frac{L}{R}$, so in our case $5\tau=\qty{187.5}{\us}$ which is longer than our current controller timing of $T_\text{ctrl,cur}=\qty{100}{\us}$ and therefore not neglectable. A second EMA was applied to the measured absolute $I_\text{raw}$, functioning as a \qty{15}{\kHz} low pass to filter differences in the current measurements caused by a different measurement timing within the PWM duty cycle. 

\subsubsection{Grasp Execution}
\label{subsection: Grasp Execution}
The software architecture implemented within our framework facilitated the initiation of predefined grasping types via the main controller software. These predefined grasps were delineated through the specification of two critical positions: a preparatory position and a target position, alongside a speed factor for each finger that orchestrated their relative movement dynamics. To activate a grasp, two interfaces with the main controller were established: a serial console and a USART control interface. In addition to specifying the grasp to be executed, these interfaces enabled the adjustment of an additional speed factor, which governed the overall velocity of the grasping action. Upon reaching the target position, the software algorithm deactivated all motor functions, leveraging the non-back-drivable nature of the worm gears to secure the grasp. This mechanism not only prevented motor overheating but also conserved energy, ensuring stable and efficient operation.

\section{Experiments}
\label{section: Experiments}
To assess the efficacy of our proposed design, we conducted a series of experiments. Initially, we calibrated and characterized the tactile sensors under different loading scenarios, including linear, dynamic, and quasi-static conditions. Subsequently, we evaluated the hand's dexterity and effectiveness through the Kapandji test, grasping experiments and a speed test.

\subsection{Tactile Sensor Characterization}

\begin{figure*}[t]
    \centering    
\includegraphics[width=0.9\linewidth]{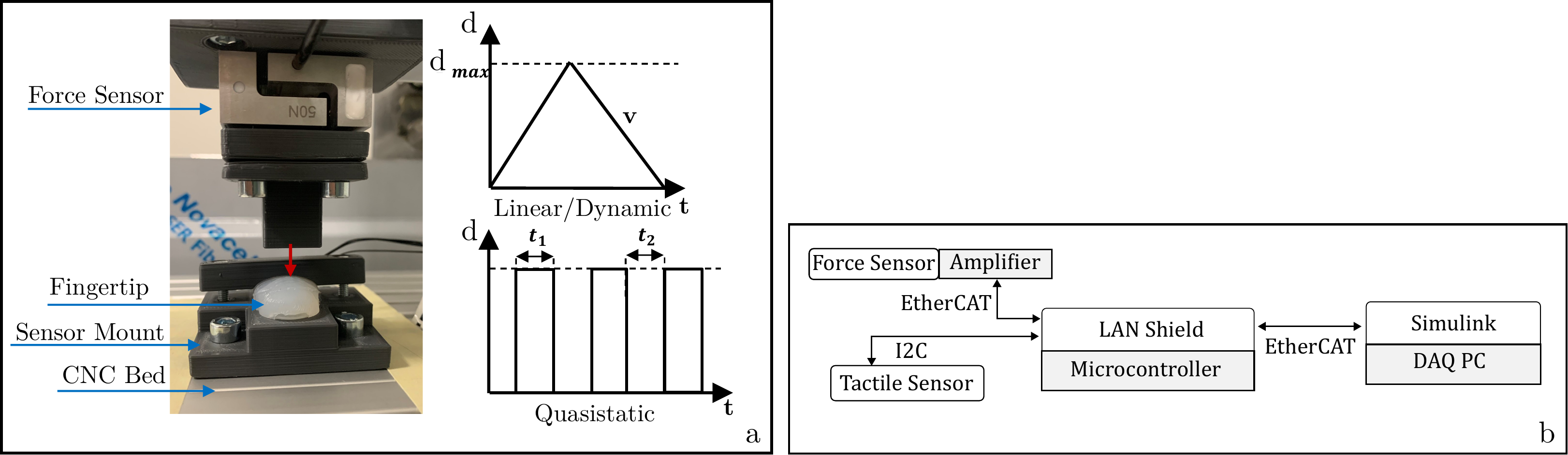}      
    \caption{Experiments for tactile sensor characterization. a) Experimental setup and procedure. b) Data acquisition.}
    \label{fig: exp_setup}  
    \vspace{-0.5\baselineskip}
\end{figure*}

For sensor characterization, we applied normal forces to the sensorized tip of each finger. As shown in \cref{fig: exp_setup}a, the displacement (d) was performed by a 4-axis CNC system (High-Z S-400T, CNC-Step GmbH, Germany) equipped with a three-axis K3D40-50N force sensor with a corresponding GSV8-DS EC/SubD44HD amplifier (ME-Messsysteme, Hennigsdorf, Germany) that served as reference. G-code commands controlled the movement of the machine to perform linear, dynamic and quasi-static loading cycles on the fingertip mounted on the CNC bed. 

A data acquisition system (DAQ) collected the sensor data using a SAME54 Xplained Pro \unit{\micro C} board (Microchip Technology Inc., USA), two custom-made PCB shields: one for connectivity with the sensors, and a real-time EtherCAT-based follower communication shield using a LAN9253 (Microchip Technology Inc., USA) controller. The \unit{\micro C} read the I$^2$C channel to collect tactile sensor data and re-transmitted it to the EtherCAT controller at \qty{1}{\kHz}.

A real-time PC (x64 running an Ubuntu 20.04) hosted the EtherCAT master using Etherlab (Ingenieurgemeinschaft IgH, Germany) at \qty{100}{\Hz}, and a post-processing environment using Matlab/Simulink (MathWorks, MA, USA). The EtherCAT master requested the sensor data of both force and tactile sensor from the \unit{\micro C} and recorded it in real-time.

Before characterization, it was necessary to determine the range (R) of each sensor, as it varied due to manufacturing inaccuracies. To achieve this, we conducted a preliminary experiment wherein the applied force was incrementally increased after initial contact with the sensor. Initially, steps of \qty{0.01}{\mm} were employed, followed by \qty{0.1}{\mm} steps of the CNC machine once the detection threshold was reached. This methodology enabled us to determine both the minimum detection threshold and the saturation point for each sensor that was subsequently applied in the characterization experiments.

For calibration, we performed 25 linear loading and unloading cycles as depicted in \cref{fig: exp_setup}a with a velocity of $v = \qty{10}{\mm\per\s}$. We calculated the initial sensor value $P_0$ and quantified the sensor drift $d$ between the first and 25th loading cycle. For calibration, we approximated the sensor's stimulus-response relationship with a transfer function $T(F_Z)$ through polynomial curve fitting (MATLAB R2022a, MathWorks, United States). We determined the sensitivity $s$ based on the linear fit for overall cycles, and the hysteresis $h$ based on $F_Z$ = R/2 during the 25th cycle. 

\begin{figure*}[tb]
     \centering
     \includegraphics[width={0.8\textwidth}]{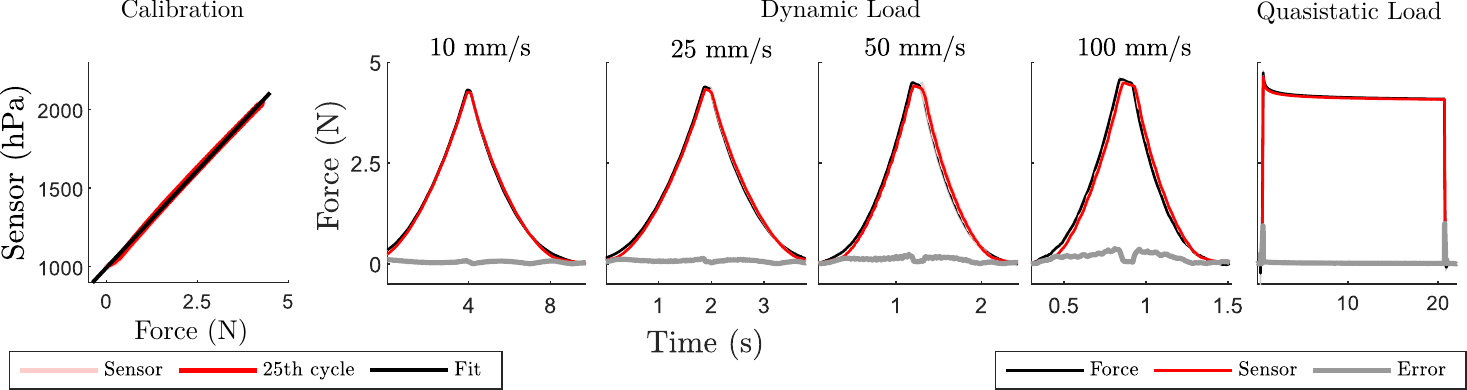}
     \caption{Results of tactile sensor characterization for the little finger sensor.}  
     \label{fig: TS_exp}
     \vspace{-1.5\baselineskip}
\end{figure*}

To evaluate the time-dependent behaviour of the sensor, we additionally performed dynamic and quasi-static experiments. In dynamic experiments, we conducted 5 cycles with increasing velocity ($v_0=\qty{10}{\mm\per\s}, v_1=\qty{25}{\mm\per\s}, v_2 = \qty{50}{\mm\per\s}, v_3=\qty{100}{\mm\per\s}$). We calculated the accuracy $\Delta F \pm \sigma$ as the mean deviation between the reference force F and estimated force F' for each case and reported the percentage with the maximum range of the sensor.

For quasi-static testing as depicted in \cref{fig: exp_setup}a ($v = \qty{300}{\mm\per\s}$), we determined the dwell times $t_1 = t_2 = \qty{20}{\s}$ in preliminary experiments as the duration for the sensor signal to reach at least 80\% of its static value after loading. We calculated the accuracy of the estimated force $\Delta F \pm \sigma$. The sensor relaxation was determined as its mechanical component ($r_m$) and sensor component ($r_s$). We calculated the mechanical relaxation as the difference of the measured ground-truth force between the beginning and end of the dwell time $t_1$ and the sensor relaxation as the difference between the estimated sensor force between the beginning and end of the dwell time.

\subsection{Grasping Experiments}

To assess the functional capabilities of the hand, we conducted a series of experiments employing predefined grasping types, each varying in target position and velocities for individual joint movements.

In the initial experiment, we evaluated thumb dexterity using the Kapandji score system \cite{KAPANDJI198667}, which assesses the ability of the thumb to make contact with different areas on the palm and fingers.   

Second, we evaluated the grasping capabilities of the hand. Therefore, we mounted the hand on a lightweight robot as depicted in \cref{fig: overview} (Panda, Franka Emika, Munich, Germany) and conducted semi-automated grasping experiments. The experiments followed the following procedure: 1) Robot movement to approach the object situated on a table, 2) manual initialization of grasping type using the USART terminal interface, 3) grasping of the object, 4) manual stop of the finger movement when the grasp was achieved, 5) robot movement to holding position, 6) holding phase of \qty{10}{\s}, 7) robot movement to shaking position, and 8) horizontal shaking motion with an S-curve trajectory, maximum velocity of \qty{0.8}{\m\per\s}, and amplitude of \qty{5}{\cm} for 5 cycles.
The robots' movements were predefined utilizing the skill framework based on \cite{johannsmeier2019framework} and joint impedance control. Cartesian impedance control was utilized for shaking movements, allowing for the implementation of an S-curve trajectory motion in a single direction. 
The Data acquisition was performed with \qty{10}{\Hz} via the USART interface. The received motor angle (\unit{\radian}), motor current (\unit{\A}), and tactile sensor data (\unit{\N}) were then processed using MATLAB (R2022a, MathWorks, United States).
The selection of grasping types and objects was based on the most prevalent grasps required in daily activities \cite{feix2014analysis, Puhlmann2016}. These encompassed Tripod, Power Sphere, Thumb-2-Finger, Lateral Pinch, Medium Wrap, Pinch Grasp, and Edge Grasp. 
For consistency with existing literature, all grasps were performed utilizing objects from the YCB object set \cite{YCB}, adhering to the AHAP protocol-specific object for each grasping type \cite{AHAP}.

Subsequently, a speed assessment was conducted by executing medium wrap movements at speed factors of 0.1, 0.5, and 1.0, with the time required for a full medium wrap recorded. Finally, the hand's resilience was evaluated by grasping a \qty{2.65}{\kg} object (a box containing wood blocks suspended on a rope from the YCB object set) and subjecting the robot to manual shaking.

%% file: Images/electronics_overview.tex
\begin{tikzpicture}[scale=\electronicsOverviewScale, every node/.style={scale=\electronicsOverviewScale}]
    \sffamily
    \tikzstyle{element}=[draw=black,fill=\electronicsOverviewColor, minimum height=0.8cm];
    \tikzset{>={Latex[length=0.9mm, width=1.6mm]}}

    \node[element, anchor=north,      minimum width=4.2cm, minimum height=2.8cm] (main)  at (0,0) {};
    
    \node[element, anchor=north west, minimum width=3.0cm, minimum height=0.8cm] (mul)   at ($(main.north east) + ( 2.0, 0.0)$) {};
    
    \node[element, anchor=north,      minimum width=3.0cm, minimum height=0.8cm] (r1)    at ($(mul.south) + (0.2, -1)$) {};
    \node[element,                    minimum width=3.0cm, minimum height=0.8cm] (r2)    at ($(r1) + (-0.1,-0.1)$) {};
    \node[element,                    minimum width=3.0cm, minimum height=0.8cm] (tact)  at ($(r2) + (-0.1,-0.1)$) {};
    
    \node[element, anchor=north east, minimum width=3.0cm, minimum height=0.8cm] (r1)    at ($(main.north west) + (-2.0, 0.)$) {};
    \node[element,                    minimum width=3.0cm, minimum height=0.8cm] (r2)    at ($(r1) + (-0.1,-0.1)$) {};
    \node[element,                    minimum width=3.0cm, minimum height=0.8cm] (motor) at ($(r2) + (-0.1,-0.1)$) {};
    
    \node[element, anchor=north,      minimum width=3.0cm, minimum height=1.8cm] (r1)    at ($(motor.south) + ( 0.2,-0.2)$) {};
    \node[element,                    minimum width=3.0cm, minimum height=1.8cm] (r2)    at ($(r1) + (-0.1,-0.1)$) {};
    \node[element,                    minimum width=3.0cm, minimum height=1.8cm] (peri)  at ($(r2) + (-0.1,-0.1)$) {};

    \draw[draw=black!40] ($(main.north west) + (-0.1,0.50)$) |- ($(main.south east) + (0.1,-0.1)$) |- ($(mul.south east) + (0.1,-0.1)$) |- ($(main.north) + (0,0.5)$) -- cycle;
    \draw[draw=black!40] ($(motor.south west) + (-0.1,-0.1)$) |- ($(motor.north east) + (0.3,0.70)$) |-  cycle;
    \node[text=black!40, anchor=south] at ($(motor.north) + (0.1,0.2)$) {Motor Board};
    \node[text=black!40, anchor=south] at ($(main.north) + (2,0)$) {Main Board};

    \node at ($(main.north) + (0,-0.4)$) {\textbf{Main Controller}};
    \node[align=left,anchor=south west]  at ($(main.south west)+(0.05,0.2)$) {- Grasp Execution\\- Tactile Sensor Readout\\- $U_\textsf{Supply}$ Monitoring\\- Data Coordination};
    
    \node at ($(mul.north) + (0,-0.4)$) {\textbf{I2C Multiplexer}};
    \node at ($(tact.north) + (0,-0.4)$) {\textbf{Tactile Sensor}};
    
    \node at ($(motor.north) + (0,-0.4)$) {\textbf{Motor Controller}};
    
    \node at ($(peri.north) + (0,-0.4)$) {\textbf{Peripherals}};
    \node[align=left,anchor=south west]  at ($(peri.south west)+(0.05,0.2)$) {- Serial Console\\- USART Control};

    \draw[<->,thick] (mul.west)   --++ (-2,0) node[pos=0.5,above]{\footnotesize BMP3 I2C};
    \draw[<->,thick] ($(motor.east) + (0.2,0)$) --++ (2,0) node[pos=0.5,above]{\footnotesize Motor I2C};
    \draw[<->,thick] ($(peri.east) + (0.2, 0.25)$) --++ (2,0) node[pos=0.5,above]{\footnotesize USART};
    \draw[<->,thick] ($(peri.east) + (0.2,-0.25)$) --++ (2,0) node[pos=0.5,above]{\footnotesize USART};
    
    \draw[<->,thick] (mul.south) --++(0,-1) node [pos=0.5,anchor=east,align=right]{\footnotesize Channel 0...7};
    
    \draw[line width=0.2mm/\electronicsOverviewScale] (-7.65, 1.7) rectangle (7.55, -4.5);

    \node[scale=1/\electronicsOverviewScale, anchor=south east] at (7.55, -4.5) {\textrm{a}};
\end{tikzpicture}

%% file: Images/control_loop.tex
\begin{tikzpicture}[scale=\controlLoopScale, every node/.style={scale=\controlLoopScale}]
    \sffamily
    \tikzstyle{element}=[draw=black, minimum height=1cm, minimum width=1cm,thick];
    \tikzstyle{elementSmall}=[draw=black, minimum height=0.4cm, minimum width=0.4cm,thick,inner sep=0];
    \tikzstyle{filter}=[draw=black, minimum height=1cm, minimum width=1cm,thick,inner sep=0];
    \tikzstyle{comp}=[circle,draw=black, minimum height=0.4cm,thick];
    \tikzstyle{mul}=[draw=black, minimum height=0.4cm, minimum width=0.4cm,thick,inner sep=0];
    \tikzstyle{dot}=[circle,draw=black,fill=black,minimum width=0.1cm, minimum height=0.1cm,inner sep=0,thick];

    \tikzset{>={Latex[length=0.9mm, width=1.6mm]}}

    \draw (0, 0)
    ++ (0,0) node (sp)                  {$\theta_\textrm{set}$}
    ++ (1.05,0) node (c0) [comp]        {}
    ++ (1.30,0) node (e0) [element]     {\textbf{\textsf{P}}}
    ++ (1.30,0) node (f0) [filter]      {\filterLimit}
    ++ (1.80,0) node (c1) [comp]        {}
    ++ (1.30,0) node (e1) [element]     {\textbf{\textsf{PID}}}
    ++ (1.30,0) node (f1) [filter]      {\filterLimit}
    ++ (1.80,0) node (c2) [comp]        {}
    ++ (1.30,0) node (e2) [element]     {\textbf{\textsf{PI}}}
    ++ (1.30,0) node (f2) [filter]      {\filterLimit}
    ++ (2.05,0) node (m)  [element]     {$M$}
    ++ (0.80,0) node (r)  [anchor=west] {$\theta_\textrm{raw}$};
    
    \draw[draw=black!40] (f0) ++ (0.75,0.75) rectangle ($(c0) + (-0.45, -3.55)$) node[pos=0.5](r1){};
    \draw[draw=black!40] (f1) ++ (0.75,0.75) rectangle ($(c1) + (-0.45, -3.55)$) node[pos=0.5](r2){};
    \draw[draw=black!40] (f2) ++ (0.75,0.75) rectangle ($(c2) + (-0.45, -3.55)$) node[pos=0.5](r3){};
    
    \draw (sp) ++ (0,1) node (l0) {$\omega_\textrm{lim}$};
    \draw (l0) ++ (0,0.4) node (l1) {$I_\textrm{lim}$};
    \draw (l1) ++ (0,0.4) node (l2) {$U_\textrm{lim}$};
    \draw[->] (l0) -| (f0);
    \draw[->] (l1) -| (f1);
    \draw[->] (l2) -| (f2);

    \draw[->,thick] (sp) -- (c0);
    \draw[->,thick] (c0) -- (e0) node[pos=0.333,anchor=south]{$\theta_\textrm{err}$};
    \draw[->,thick] (e0) -- (f0);
    \draw[->,thick] (f0) -- (c1) node[pos=0.5,anchor=south]{$\omega_\textrm{set}$};
    \draw[->,thick] (c1) -- (e1) node[pos=0.333,anchor=south]{$\omega_\textrm{err}$};
    \draw[->,thick] (e1) -- (f1);
    \draw[->,thick] (f1) -- (c2) node[pos=0.5,anchor=south]{$I_\textrm{set}$};
    \draw[->,thick] (c2) -- (e2) node[pos=0.333,anchor=south]{$I_\textrm{err}$};
    \draw[->,thick] (e2) -- (f2);
    \draw[->,thick] (f2) -- (m) node[pos=0.6,anchor=south]{$U_\textrm{set}$} node[dot,pos=0.55] (d1) {};
    \draw[->,thick] (m)  -- (r);

    \draw (f2) ++ (0,-1.5) node (f3) [filter] {\filterEMA};
    \draw (f3) ++ (-1.3,0) node (f4) [filter] {\filterSign};
    \draw (c2) ++ (0,-1.5) node (mul)[mul]{\textbf{\textsf{\texttimes}}};
    \draw[->,thick] (d1) |- (f3);
    \draw[->,thick] (f3) -- (f4);
    \draw[->,thick] (f4) -- (mul);
    
    \draw (f3) ++ (0,-1.3) node (f5) [filter] {\filterEMA};
    \draw[->,thick] ($(m.south) + (-1/6,0)$) |- (f5);
    \draw[->,thick] (f5) -| (mul);
    \draw[->,thick] (mul) -- (c2) node[pos=1,anchor=north west] {\textbf{\textsf{\Large-}}};

    \draw[->,thick] (f1) --++ (0,-0.75) -| (e1);
    \draw[->,thick] (f2) --++ (0,-0.75) -| (e2);

    \draw[thick] ($(m.south) + (1/6,0)$) |-  ($(c1) + (0,-3.80)$) node (d4) [pos=1,dot]{};
    \draw(c0) ++ (0,-1.5) node (s) [elementSmall] {\scriptsize\textbf{\textsf{s\textsuperscript{-1}}}};
    \draw[->,thick] (d4) -| (s) -- (c0) node[pos=1,anchor=north west]{\textbf{\textsf{\Large-}}};
    \draw[->,thick] (d4) -- (c1) node[pos=1,anchor=north west]{\textbf{\textsf{\Large-}}};

    \node[anchor=south east] at($(m.south) + (-1/7,-0.5)$) {$I_\textrm{raw}$};
    \node[anchor=south west] at($(m.south) + (1/6,-0.5)$) {$\omega_\textrm{raw}$};
    
    \node at ($(r1) + (0,-1.8)$) {\textcolor{black!40}{Position}};
    \node at ($(r2) + (0,-1.8)$) {\textcolor{black!40}{Velocity}};
    \node at ($(r3) + (-0.5,-1.8)$) {\textcolor{black!40}{Current}};
    
    \draw[line width=0.2mm/\controlLoopScale] ($(sp.west) + (-0.25, 2.15)$) rectangle ($(r.east) + (0.25, -4.05)$);
    
    \node[scale=1/\controlLoopScale, anchor=south east] at ($(r.east) + (0.25, -4.05)$) {\textrm{b}};
\end{tikzpicture}

%% file: Sections/Results.tex
\section{RESULTS}
\label{section: results}
\input{Tables/SensorChar}

\subsection{Sensorized Fingertips}
\subsubsection{Preliminary experiments}
\begin{figure}[tb]
    \centering
\includegraphics[width=0.9\linewidth]{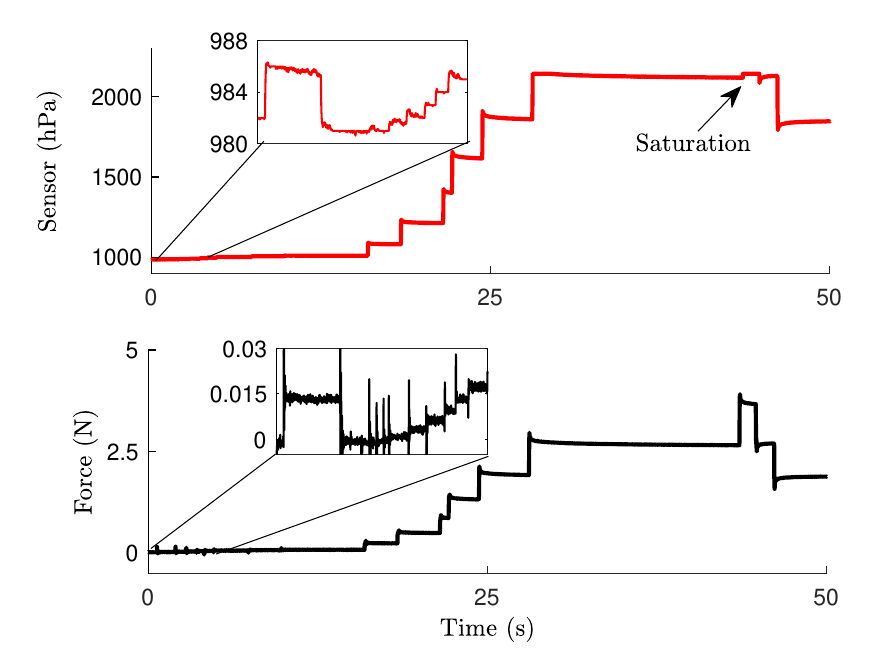}
    \caption{Results of pre-experiments for thumb sensor.}  
    \label{fig:pre_exp}
    \vspace{-1.5\baselineskip}
\end{figure}

We conducted preliminary experiments for 5 sensorized fingertips each resulting in a different force range. 
\Cref{fig:pre_exp} depicts the result for the thumb sensor. The sensor showed saturation at an applied force above \qty{2.32}{\N}, while the other sensors showed ranges between \qty{3.66}{\N} and \qty{9.46}{\N} (see \cref{tab: TS_exp}). \Cref{fig:pre_exp} also depicts the incremental increase in force at the beginning of the experiment. It shows, that the minimal detectable change in sensor signal was below an applied force of \qty{0.01}{\N}.

\subsubsection{Sensor Characterization}

Sensor characteristics for all experiments are displayed in \cref{tab: TS_exp}. \Cref{fig: TS_exp} shows the calibration results of the little finger sensorized fingertip (refer to the attached repository for other fingers). We used linear fits for the little, ring and middle finger while the index and thumb showed better results with a 2nd-order polynomial. The displayed fit functions led to $R^2 > \qty{99}{\percent}$ for all prototypes. The sensors showed a drift over 25 cycles between \qty{0.2}{\percent} and \qty{1}{\percent} of their zero value. The zero value varied between \qty{973}{\hecto\Pa} for the ring finger and \qty{1008}{\hecto\Pa} for the index finger. The sensitivity ranged between \qty{201}{\hecto\Pa\per\N} to \qty{298}{\hecto\Pa\per\N} for the little, ring and middle finger, while the index finger showed a lower sensitivity with \qty{103}{\hecto\Pa\per\N} and the thumb showed a larger sensitivity with \qty{462.08}{\hecto\Pa\per\N}. All sensors showed a hysteresis \qty{<3}{\percent}. 

The dynamic experiments revealed that the sensor accuracy decreased with increased velocities. An example is visible in the resulting graphs of the little finger displayed in \cref{fig: TS_exp}. With higher velocities, the sensor curve showed an increasingly delayed response. This is reflected in the accuracy for all sensors (see \cref{tab: TS_exp}). While it ranged between $\Delta F_{Thumb}$ = \qty{1.06 +- 0.54}{\percent} to \qty{1.49 +- 0.78}{\percent} for a velocity of \qty{10}{\mm\per\s} it increased up to $\Delta F_{Index}$ = \qty{3.69 +- 3.28}{\percent} for higher velocities. During quasi-static experiments, all sensors showed mechanical and sensor relaxation. The mechanical relaxation showed higher values between $r_{m}$ = \qty{13.12}{\percent} to \qty{37.99}{\percent} while the sensor relaxation ranged between $r_{s}$ = \qty{7.51}{\percent} to \qty{15.05}{\percent}. As depicted in \cref{fig: TS_exp} the relaxation leads to estimation errors in the loading/ unloading phase up to 1N for the little finger sensor.

\begin{figure*}[tb]
    \centering
\includegraphics[width={0.95\textwidth}]{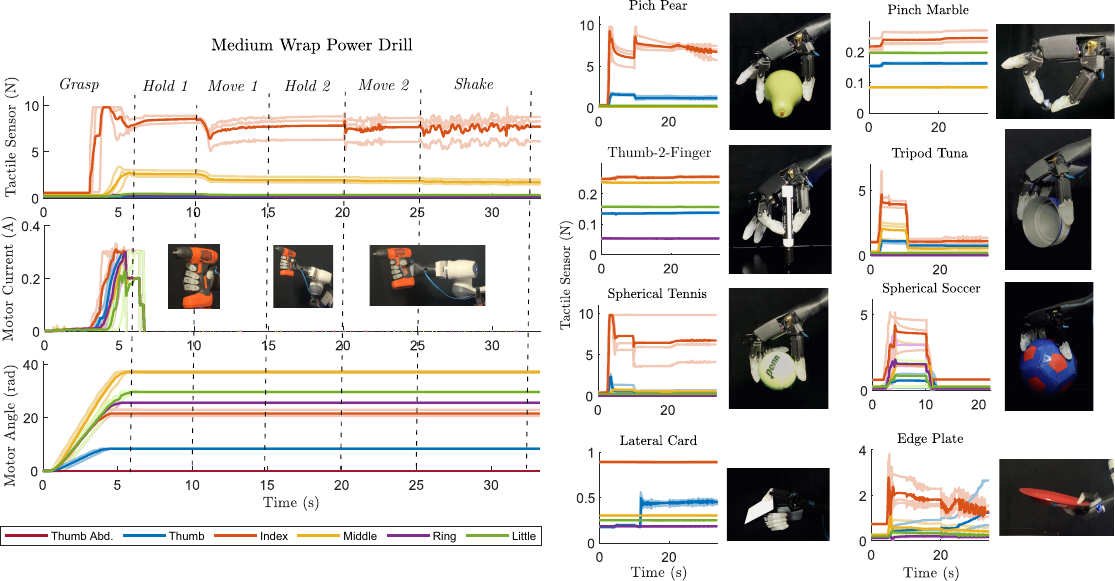}
    \caption{Results for grasping experiments showing the results from 3 repetitions (light colours) and their mean (dark colours).}
    \label{fig: exp1}
    \vspace{-1.5\baselineskip}
\end{figure*}
\subsection{Grasping Experiments}
During the Kapandji test, the thumb successfully reached positions one to four situated on the index and middle finger. However, due to the limited workspace resulting from its maximum abduction, points 5-10 on the other fingers and palm were inaccessible.

We conducted grasping experiments with 12 objects and 7 grasping types and reported the tactile sensor data, motor current for each active joint, and motor angle. As exemplified in \cref{fig: exp1}, the experimental results for the medium wrap experiment with the power drill object are presented, showcasing distinct phases of the experiment. During the grasping phase, all motor angles increased until contact was made with the drill's surface. This contact event was reflected in both the tactile sensor data and motor current. Motor currents increased uniformly up to a value of \qty{0.3}{\A} (control threshold), with the index finger exhibiting the fastest increase and the little finger the slowest. Correspondingly, tactile sensor data indicated a force increase on the index finger, followed by the middle and little fingers with lower forces. Despite indications from the motor current of contact between the thumb and ring finger, no increase in tactile sensor signal was observed. Before the index motor reaches its final position, the tactile force decreases from \qty{10}{\N} to approximately \qty{8}{\N}, then increases again to around \qty{9}{\N} toward the initial holding phase.
After reaching the final grasping position, the motors were stopped, relying solely on the non-backdrivable behaviour of the actuation train to maintain the grasping force. Consequently, motor currents returned to \qty{0}{\A}, and motor angles remained constant throughout the remainder of the experiment.
Subsequently, the robot transitioned to the second holding position, resulting in a noticeable change and vibration in the tactile sensor signal. Throughout the second holding phase, the tactile data remained constant. Finally, during shaking, there was a discernible change in tactile data, indicating that the grip was retained.

\Cref{fig: exp1} provides an overview of comparable experiments with eight other objects. The figures showcase the grasped objects and corresponding tactile sensor data. The three remaining grasping experiments, motor current and position data can be found in the referred repository. Except for one instance involving the spherical grasp of a mini soccer ball, all experiments were successfully conducted, signifying stable grasps throughout the aforementioned phases. In the case of the soccer ball, the experimental data indicated that the ball could not be lifted from the table and was dropped during the first robot movement, resulting in a sudden decrease in tactile data after \qty{10}{\s}. 

During the experiments, certain tactile sensors exhibited offsets, particularly noticeable in the tripod marker experiment. In this experiment only small forces were applied, which did not significantly deviate from the initial sensor offset in the index finger of approximately \qty{0.3}{\N}. Moreover, in the lateral pinch experiment, when gripping the credit card, it tilted sideways due to the grasping surface being solely on the edge of the palm. In several instances, particularly during spherical grasps, we observed the thumb bending backwards under the force exerted by other fingers after initial contact, preventing it from fully wrapping around the object.

The speed test results indicated that the hand could achieve minimum closing times of \qty{8}{\s} and \qty{6}{\s}, and opening times of \qty{6}{\s} and \qty{3}{\s} for speed factors 0.1 and 0.5, respectively. For the highest speed factor of 1.0, the closing time decreased to \qty{3}{\s} while the opening time was \qty{2}{\s}.
Additionally, the hand successfully grasped and endured shaking movements while holding the \qty{2.65}{\kg} block box. For visual demonstrations of the Kapandji, speed, and block box experiments, please refer to the attached video and repository.

%% file: Tables/SensorChar.tex
 \bgroup
 \def\arraystretch{1.2}
 \setlength\tabcolsep{0.5em}
\begin{table*}[tb]
\centering
    \caption{Sensor characteristics for calibration, dynamic, and quasistatic loading experiments} R: Range (N). $P_0$: Zero value (\unit{\hecto\Pa}). s: sensitivity (\unit{\hecto\Pa\per\N}). h: hysteresis (\unit{\N}) between loading and unloading at half the Range of the 25th cycle. $\Delta F \pm \sigma$: Accuracy (\unit{\percent}) as mean error and standard deviation of calibrated signal in relation to sensor range. $r_m$: Mechanical relaxation (\unit{\percent}) indicated by decreased force between beginning and end of dwell time in relation to range. $r_{s}$: Sensor relaxation (\unit{\percent}) indicated by the decrease in detected force between beginning and end of dwell time in relation to range.
\begin{tabular}{|l|llll|llllllll|llll|}
\hline
\multicolumn{1}{|c|}{\multirow{2}{*}{\textbf{Experiment}}} & \multicolumn{4}{c|}{\multirow{2}{*}{\textbf{Calibration}}}                                                                         & \multicolumn{8}{c|}{\textbf{Dynamic}}                                                                                                                                                                                                & \multicolumn{4}{c|}{\multirow{2}{*}{\textbf{Quasistatic}}}                                        \\ \cline{6-13}
\multicolumn{1}{|c|}{}                                     & \multicolumn{4}{c|}{}                                                                                                              & \multicolumn{2}{c|}{\textbf{v=10mm/s}}                       & \multicolumn{2}{c|}{\textbf{v=25mm/s}}                       & \multicolumn{2}{c|}{\textbf{v=50mm/s}}                       & \multicolumn{2}{c|}{\textbf{v=100mm/s}} & \multicolumn{4}{c|}{}                                                                             \\ \hline
Characteristics    & \multicolumn{1}{l|}{R} & \multicolumn{1}{l|}{$P_0$} & \multicolumn{1}{l|}{s} & \multicolumn{1}{l|}{h} & \multicolumn{1}{l|}{$\Delta F$} & \multicolumn{1}{l|}{$\pm \sigma$}  & \multicolumn{1}{l|}{$\Delta F$} & \multicolumn{1}{l|} {$\pm \sigma$}  & \multicolumn{1}{l|}{$\Delta F$} & \multicolumn{1}{l|}{$\pm \sigma$}  & \multicolumn{1}{l|}{$\Delta F$}   & \multicolumn{1}{l|}{$\pm \sigma$}    & \multicolumn{1}{l|}{$\Delta F$} & \multicolumn{1}{l|}{$\pm \sigma$} & \multicolumn{1}{l|}{$r_m$} & $r_s$ \\ \hline
Sensor Pinky                                               & \multicolumn{1}{l|}{4.30}      & \multicolumn{1}{l|}{995.00}           & \multicolumn{1}{l|}{246.97}              & 2.16           & \multicolumn{1}{l|}{1.06}        & \multicolumn{1}{l|}{0.54} & \multicolumn{1}{l|}{1.70}        & \multicolumn{1}{l|}{0.63} & \multicolumn{1}{l|}{2.27}        & \multicolumn{1}{l|}{1.33} & \multicolumn{1}{l|}{3.67}      & 2.48   & \multicolumn{1}{l|}{0.63}   & \multicolumn{1}{l|}{1.65}    & \multicolumn{1}{l|}{14.43}   & 12.68 \\ \hline
Sensor Ring                                                & \multicolumn{1}{l|}{4.57}      & \multicolumn{1}{l|}{973.00}           & \multicolumn{1}{l|}{201.54}              & 2.28           & \multicolumn{1}{l|}{1.41}        & \multicolumn{1}{l|}{0.91} & \multicolumn{1}{l|}{1.97}        & \multicolumn{1}{l|}{0.91} & \multicolumn{1}{l|}{2.33}        & \multicolumn{1}{l|}{1.46} & \multicolumn{1}{l|}{3.59}      & 2.58   & \multicolumn{1}{l|}{1.79}   & \multicolumn{1}{l|}{1.50}    & \multicolumn{1}{l|}{15.58}   & 12.46 \\ \hline
Sensor Middle                                              & \multicolumn{1}{l|}{3.66}      & \multicolumn{1}{l|}{989.00}           & \multicolumn{1}{l|}{298.75}              & 2.83           & \multicolumn{1}{l|}{1.36}        & \multicolumn{1}{l|}{0.86} & \multicolumn{1}{l|}{2.04}        & \multicolumn{1}{l|}{1.04} & \multicolumn{1}{l|}{2.53}        & \multicolumn{1}{l|}{1.57} & \multicolumn{1}{l|}{3.99}      & 2.51   & \multicolumn{1}{l|}{1.29}   & \multicolumn{1}{l|}{1.63}    & \multicolumn{1}{l|}{15.20}   & 12.24 \\ \hline
Sensor Index                                               & \multicolumn{1}{l|}{9.46}      & \multicolumn{1}{l|}{1008.00}          & \multicolumn{1}{l|}{103.47}              & 2.96           & \multicolumn{1}{l|}{1.49}        & \multicolumn{1}{l|}{0.78} & \multicolumn{1}{l|}{2.00}        & \multicolumn{1}{l|}{0.94} & \multicolumn{1}{l|}{2.45}        & \multicolumn{1}{l|}{1.43} & \multicolumn{1}{l|}{3.89}      & 2.26   & \multicolumn{1}{l|}{0.96}   & \multicolumn{1}{l|}{1.54}    & \multicolumn{1}{l|}{13.12}   & 7.51  \\ \hline
Sensor Thumb                                               & \multicolumn{1}{l|}{2.32}      & \multicolumn{1}{l|}{974.00}           & \multicolumn{1}{l|}{462.08}              & 1.75           & \multicolumn{1}{l|}{1.26}        & \multicolumn{1}{l|}{0.76} & \multicolumn{1}{l|}{1.59}        & \multicolumn{1}{l|}{1.08} & \multicolumn{1}{l|}{2.44}        & \multicolumn{1}{l|}{1.90} & \multicolumn{1}{l|}{3.69}      & 3.28   & \multicolumn{1}{l|}{1.64}   & \multicolumn{1}{l|}{1.63}    & \multicolumn{1}{l|}{37.99}   & 15.05 \\ \hline
\end{tabular}
\label{tab: TS_exp}
\end{table*}

%% file: Sections/Discussion_and_conclusion.tex
\section{DISCUSSION AND CONCLUSION}
\label{section: discussion and conclusion}
The preliminary experiments and grasping tests with sensorized fingertips offer important insights into the capabilities and areas for improvement of the hand. The initial findings highlight the hand's sensitivity to a wide range of forces, though variations in sensor performance across different fingers suggest the need for improved manufacturing. The experiments successfully demonstrated the hand's versatility in grasping diverse objects, showcasing its adaptability. However, challenges such as the thumb's limited range and occasional backward bending under force highlight specific design limitations that could impact the efficacy of the grasp. Furthermore, the hand's ability to maintain stable grasps, even during dynamic movements, emphasizes its potential utility. Still, instances of sensor offset and difficulty in grasping large objects point to the necessity for ongoing refinement in sensor integration and mechanical design. These results underscore a solid foundation for future enhancements, aiming to optimize the hand's performance across a broader range of real-world applications.

\subsection{Hand Design}
We developed a robotic hand with an integrated 6 DOF actuation system, corresponding electronics and control, tactile sensitivity, and a compliant linkage mechanism, all contained within a human-sized palm. The hand showed sufficient grasping capabilities with 11 objects utilizing 6 grasping types, a minimum closing speed of \qty{3}{\second} and a maximum tested payload of \qty{2.65}{\kg}.
Despite these features, the hand remains cost-effective, with a total cost of approximately 500\,EUR. This affordability places it in a favourable position compared to similar solutions, as demonstrated in \cref{tab: robotic_hand_comparison}. Additionally, we have made the design open-source, providing accessibility to researchers and developers for future advancements.
Our implementation of the compliant linkage mechanism has proven effective during 11 out of 12 grasping experiments, allowing the fingers to adapt to various object shapes and holding it during the shaking phase solely depending on the loaded spring and non-backdrivable actuation unit. This is comparable to other mechanisms as proposed by \cite{choi2017compliant}.
We have adapted the compliant linkage concept proposed by \cite{choi2017compliant}, by utilizing rotational joints with rotational and extension springs to achieve compliance in the fingers with off-the-shelf parts.
This came at the cost of robustness but minimized sideways movement of the distal phalanx, enhancing control over finger movements. Future work will focus on increasing the robustness of the hand, exploring alternative materials and reinforcing parts subjected to high loads, such as motor holders and proximal linkages.
The Kapandji test showed the need to enhance thumb dexterity and workspace to enable more advanced manipulation. During grasping experiments, we encountered difficulties with stable grasp of large objects particularly when the distal links failed to close further due to object contact with the proximal links. 

\subsection{Electronics}

We developed custom PCBs and a software framework that enabled current, velocity and position control comparable to other solutions as proposed by \cite{Edmundo2023}.
In our future work, we aim to implement current-induced contact detection and sensor fusion with tactile data to enable a seamless transition to current control upon contact detection by all fingers. This sets the baseline for automatic equal force distribution on grasped objects. 
Further improvements will include an improved USART control interface featuring individual low-level control of all fingers, improved data transmission rates and the development of a motor temperature model. This will prevent overheating while allowing the increase in the used maximum current and therefore motor torques. 

\subsection{Tactile Sensors}
We have successfully integrated tactile sensors into the fingertips of our robotic hand, achieving high sensitivity with a threshold below \qty{0.01}{\N}, accuracy between \qty{1.06 +- 0.54}{\percent} and \qty{3.99 +- 2.51}{\percent} in static and dynamic conditions, hysteresis below \qty{3}{\percent}, and a sensor relaxation below the mechanical relaxation of the silicone. These results show advances or are comparable to similar studies utilizing barometer-based principles \cite{Koiva2020, Jentoft2013, Al2023}. Furthermore, our design and manufacturing process enabled the robust bonding of the SLA printed distal phalanx with the silicone without the need for additional adhesives which was not implemented in comparable literature \cite{Koiva2020}.
Future work will focus on implementing additional modalities such as slippage and shear force as proposed by \cite{Kim2023, Clercq2022}. 
The characterization and grasping experiments revealed different force ranges for the sensor prototypes and an observed sensor offset up to approx. 0.5\,N as well as saturation at high loads. Future research will focus on an improved manufacturing process to generate consistent force ranges and saturation, and further analysis of the sensor offset.

In summary, we provided a robotic hand design including a compliant linkage mechanism with 6 DOA, tactile sensing capabilities, the corresponding electronics and control all integrated into a human-sized prototype. The experiments and grasping tests with sensorized fingertips offer important insights into the capabilities and areas for improvement of the hand. The initial findings highlight the hand's sensitivity to a wide range of forces, though variations in sensor performance across different fingers suggest the need for improved manufacturing. The experiments successfully demonstrated the hand's versatility in grasping diverse objects, showcasing its adaptability. However, challenges such as the thumb's limited range and occasional backward bending under force highlight specific design limitations that could impact the efficacy of the grasp. Furthermore, the hand's ability to maintain stable grasps, even during dynamic movements, emphasizes its potential utility. Still, instances of sensor offset and difficulty in grasping large objects point to the necessity for ongoing refinement in sensor integration and mechanical design. These results underscore a solid foundation for future enhancements, aiming to optimize the hand's performance across a broader range of real-world applications.

